# Understanding the Impact of News Articles on the Movement of Market Index: A Case on Nifty 50

Subhasis Dasgupta
subhasisdasgupta1@acm.org
Praxis Business School, Kolkata

Pratik Satpati
pratik05062021@gmail.com
Praxis Tech School, Kolkata

Ishika Choudhary
raiishika15@gmail.com
Praxis Tech School, Kolkata

Jaydip Sen
jaydip.sen@acm.org
Praxis Business School, Kolkata

*Abstract-* In the recent past, there were several works on the prediction of stock price using different methods. Sentiment analysis of news and tweets and relating them to the movement of stock prices have already been explored. But, when we talk about the news, there can be several topics such as politics, markets, sports etc. It was observed that most of the prior analyses dealt with news or comments associated with particular stock prices only or the researchers dealt with overall sentiment scores only. However, it is quite possible that different topics having different levels of impact on the movement of the stock price or an index. The current study focused on bridging this gap by analysing the movement of Nifty 50 index with respect to the sentiments associated with news items related to various different topic such as sports, politics, markets etc. The study established that sentiment scores of news items of different other topics also have a significant impact on the movement of the index.

## I. INTRODUCTION

It is a known fact that if a stock market's movements can be predicted, great fortune can be accumulated within a very short period. However, the drivers of stock market are quite large in number, and it is governed significantly by the investors' sentiments [1]. However, it was also seen by researchers that when the overall market sentiments are not getting affected by any sudden stimulus (e.g. war or bankruptcy of a large organization etc.), the stock market's movements can be predicted by their recent past data [2]. This process is effective for a stock which is less volatile in nature. However, for a more volatile stock, the predictions suffer a lot due to investors' sentiment. In univariate models, external variables do not play any role. But it is to be understood that stock market is driven by many different factors at both micro and macroeconomic levels. Hence, along with its recent past data, others influential factors are also required to be included in the model for a better predictive accuracy. There are researches available where the researchers used sentiment scores to predict stock prices [3], [4], [5]. However, a gap was observed in this area of research as the researchers used the overall sentiment score as predictor apart from the past data of the same stock price. The current project tried to bridge the gap with topic specific sentiments. The objective of the study was to predict the index movement for one day in future.

The work was not really meant for predicting the index value in future. Rather, it was done on NIFTY 50 index to understand the impact of news sentiments of various news items on the movement of the index. And to the best of the knowledge of the researchers, no such prior work was done on movement of this index based on several news topics. Hence, a comparative analysis was not possible for this work.

The rest of the paper is organized as follows. Section II deals with the related work where some previous works are highlighted. Section III discusses the methodology at length and section IV deals with the analysis and subsequent results. Finally, section V concludes the work with conclusive remarks.

## II. RELATED WORK

In the past, there were significant numbers of research work in stock price predictions. The researchers used univariate sequence modelling such as ARIMA, Hidden Markov Model (HMM) [6], LSTM [7] and GRU [8]. In fact, ensemble models were also used for predicting the stock prices. There were researchers who attempted to solve this problem of predicting stock prices using financial metrics such as sharp ratio and GARCH model [9], [10] and sharp ratio with deep learning models [10]. There are also research works in this area using reinforcement learning [11]. Recently, the same problem was tried to be solved after incorporating sentiment scores associated with the news or reviews available in social media or newspapers [12], [13], [14]. However, there is a lack of research contribution in this area when topic-based sentiment analysis is required to be used for predicting stock price or index movements. The current study tried to investigate if apart from market related news is there any other type of news that can have a significant impact on the stock price movement.

## III. METHODOLOGY

To conduct the study, a significant number of news items were required to be collected first. To collect the news items, the archive of a popular Indian newspaper was chosen. The newspaper was Economic Times. During the COVID pandemic, stock market all over the world got severely affected. From the year 2021, India started recovering from the grip of the pandemic. Hence, for this study, data were collected from Jan 2021 till 22nd Feb 2024. The news data were scrapped from the archive of the Economic Times using a python package BeautifulSoup. Total number of news scraped was above 4 lakhs. While gathering data, due attention was given to make sure that the scrapping process could not load the host server due to repeated pings within a very short duration of time. In this study, not an individual stock price was predicted. Rather, market index of NIFTY 50 was chosen for prediction within the same period. For extracting the topics, no modelling technique was used. The links of the news items were created in such a way that the topics were mentioned in the new links only. For example, if a news was belonging to politics, the news link contained the word politics in it and using regular expression, the topic was extracted quite easily from the link itself. Not only that, but each news item was also accompanying with the news headline and hence, while scraping the data, dates, headlines, news links and the actual news were extracted and saved as pickle files. The topics were extracted from the news links and attached to the existing dataset. Afterwards, sentiment analysis was performed. For sentiment analysis, two separate methods were employed as described below.

*(1) Sentiment analysis with VADER*

Sentiment analysis with VADER [15] is a sentiment lexicon-based analysis. Here, the text is broken down to its individual words and for each word the sentiment score is extracted. The scores range between -4 to +4 which are normalized to lie between -1 to +1. A +1 scores denotes the highest positive sentiment and a -1 denotes a lowest negative sentiment. Finally, a composite score is calculated using the formula,

$$s = \frac{(\sum word\ scores)}{\sqrt{(\sum word\ scores^2 + \alpha)}} \quad (1)$$

where, $\sum word\ score$ denotes the sum of all individual word's sentiment score and $\alpha$ is a smoothing parameter which typical value is 15. Sentiment analysis with VADER is quite simple and fast. For extracting the overall sentiment of a news, this method is quite applicable as the number of words used in a proper news coverage can go easily beyond 500 words and VADER sentiment analyser works well in such scenarios. But the same was not used for finding sentiment scores associated with the news headlines.

*(2) Sentiment analysis with DistilBERT*

After the transformer model proposed by Vaswani et al. [16], the first most prominent model came to existence was BERT [17] which stands for Bidirectional Encoder Representation from Transformers. BERT was a model for serious Natural Language Understanding task such as summarization, question and answer system, text classification etc. However, the number of parameters in a BERT model was 110 million for the base model and for the large model, the number was 340 million. DistilBERT [18] is a lighter version of BERT with number of parameters being 66 million, much lesser than that of BERT. However, the model's performance was close to BERT (~97% of BERT's performance). Due to the lighter version, DistilBERT runs faster than BERT and for text classification task, there is not a significant drop of performance compared to the original BERT model. Hence, it was deemed fit for extracting sentiment scores of headlines. VADER sentiment analyser always gives a neutral sentiment when the word under consideration is not a part of the sentiment lexicon. For example, "Humanitarian act" is positive in sense and any such act should have positive sentiment. But VADER sentiment analyser gives a neutral sentiment. But DistilBERT model gives a strong positive sentiment because it was trained on 16 GB BERT data (3.3 billion words). Hence, understanding the context and is much better for DistilBERT and hence the classification efficacy. It is quite natural to assume that if the sentiment of the news headline is positive, the sentiment of the corresponding news would also be positive. Hence, extracting sentiment scores of headlines was considered appropriate for the analysis. DistilBERT gives binary outcome, i.e., POSITIVE and NEGATIVE along with the probability score of the predicted class. For the current study, the requirement was a numeric score. Hence, the probabilities were used to extract the corresponding regression scores. For example, if the predicted class is POSITIVE with probability 98%, then the regression score is $\log\left(\frac{0.98}{1-0.98}\right) = 1.69$ and if the predicted class is NEGATIVE with the probability score of 98%, the regression score is $\log\left(\frac{1-0.98}{0.98}\right) = -1.69$. DistilBERT is slower than VADER even with NVIDIA T4 GPU available in Google Colab. Hence, this method was not used for extracting sentiment scores of the news as the text length for the news were much higher than that of the headlines and it would have taken much longer time to analyse the news and give the sentiment score by DistilBERT.

IV. RESULTS AND ANALYSIS

The web scrapping part of this study was a significant task as all the news articles of each day were collected for more than three years. The total number of data points collected was above 400k and it took nearly 20 days to scrape the data in a way so that the webserver was never overloaded due to repeated requests sent within a small duration of time. After the collecting the data, topics were extracted from the URL links of the news articles. The number of such topics were 53 but there were many topics available which were having very low frequency of occurrence. The threshold frequency was taken as 200 and all the topics which were below this threshold were neglected. Thus, finally, there were 22 topics left for further analysis. In the processed dataset, for each headline and the corresponding news, there was the associated topic and the associated data of publication. Next, for news, VADER sentiment analysis was used to extract the overall sentiment of each news item and DistilBERT was used to extract the sentiment score of each headline. The sentiment scores were added to the existing dataset and the resulted tall dataset was reshaped to a wide dataset where each row had a unique date and there were 22 columns having the topics and the sentiment scores were there at the intersection of a date and a topic. Missing values in the sentiment scores were replaced by 0. In fact, two such datasets were produced. One with sentiment scores of news and the second was with sentiment scores of news headlines. The dataset of NIFTY 50 data from 1st Jan 2021 till 22nd Feb 2024 were also downloaded. This dataset contained "Open", "High", "Low" and "Close" values of the NIFTY 50 index date wise. The NIFTY 50 dataset contained periodic missing values because stock market stays close on Saturdays and Sundays. But it was quite natural to assume that news of Saturdays and Sundays can easily influence the movement of the Index on Monday. That is why, the news sentiment dataset was left joined with the NIFTY 50 dataset so that sentiment scores of the Saturdays and Sundays were not removed during joining two datasets. The missing values introduced due to left joining in the variables "Open", "High", "Low" and "Close" were filled with forward fill process (i.e., last value carry forward). This way two complete set of datasets were created with two different sets of sentiment scores. The "Close" variable was considered as the target variable. Next, for each dataset, two different lags were considered, i.e., up to 3 lags and up to 5 lags. It was assumed that movement of NIFTY 50 index was dependent of immediate past news and hence, these two separate lag values were considered in the analysis. Thus, in total, 4 datasets were created. Total number of data points in each dataset was above 1000 and the datasets were split into training and test datasets. The training dataset was comprising of data points between 1st Jan 2021 till 31st Aug 2023. The test dataset was comprising of data points between 1st Oct 2023 to 22nd Feb 2024. Web scraping was hampered for several days in the month of September 2023 due to which the data for the whole month was discarded. With up to 3 lags, the number of columns increased from 26 to 78 and with up to 5 lags, the number of columns increased from 26 to 130. Since the number of columns was more, affine function based regression analysis was considered for this study. First, multiple regression analysis was conducted with the lagged values of the available columns. After training the models, they were applied on the corresponding test data. The prediction of the models vis-à-vis the actual data are shown in Figure 1 and Figure 2 for lag 3 and lag 5 respectively. The prediction was matching the actual index movement quite closely. The RMSE values are also mentioned in the corresponding graphs. However, for comparative study, it was necessary to check the RMSE scores for a model which was trained on only the lagged values of "Open", "High", "Low" and "Close". The RMSE observed was 134.85.

That way, with VADER sentiment scores on the news data, a marginal improvement in RMSE was observed with up to lag 3 data. But the performance went down with up to lag 5 data. Linear regression has several important assumptions related to the distribution of errors and multicollinearity. Even though the results were impressive, the condition number was very high and the error terms were significantly deviated from the normal distribution. Hence, regularized regression algorithms were used. Two such commonly used algorithms are Ridge and Lasso regression.

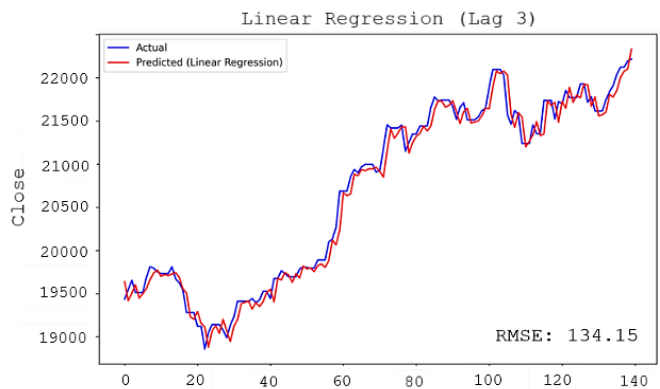

**Figure 1**. Performance of linear regression model on test data with lagged data up to lag 3

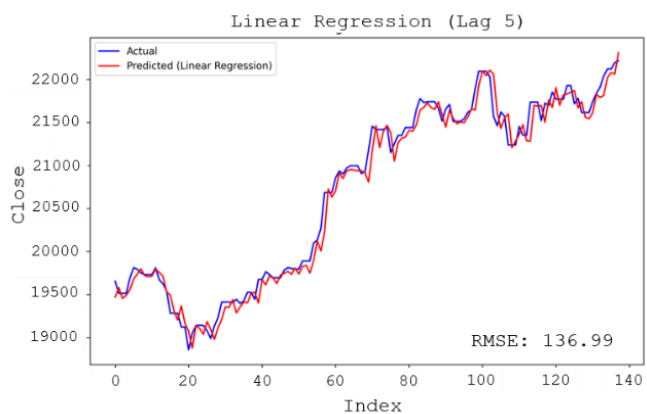

**Figure 2.** Performance of linear regression model on test data with lagged data up to lag 5

The comparative analysis is shown in Table 1. It can be seen that regularized regression scores, after fine tuning, produced better results with sentiment scores than that of without sentiment scores. Lasso regression performed very badly as it brought down the coefficients of many variables to 0. Ridge regression, on the other hand, brought down the coefficients of the less important variables to a small value but did not make them zero. This helped the model to perform in a much better way than Lasso regression. Lasso and Ridge regression are two extremes and Elastic Net is a model that lies in between the two extremes. Performance of Elastic Net was found best for the dataset without any sentiment score.

The next analysis involved the similar steps that was done earlier but the sentiment scores were based on news headlines using DistilBERT model. With DistilBERT, the understanding of the context and the associated sentiments were extracted more accurately and that helped in achieving better performance in prediction. Table 3 shows the performance of the predictive models using sentiment scores extracted by DistilBERT. It can be seen that there is a significant drop in the RMSE score of both Ridge and Lasso regression when these sentiment scores were used.

**Table 1**. Performance comparison of models with and without sentiment scores of news using VADER sentiment scores

| Model | RMSE (Without sentiment scores (lag 3)) | RMSE (With sentiment scores (lag 3)) | RMSE (With sentiment scores (lag 5)) |
|---|---|---|---|
| Linear Regression | 134.85 | **134.15** | 136.99 |
| Ridge Regression | 203.78 | **180.7** | 186.10 |
| Lasso Regression | 3367.2 | 3780.78 | 3791.10 |
| Elastic Net Regression | **174.81** | 395.56 | 392.02 |

**Table 2**. Top 5 positively contributing sentiment features in Ridge regression based on VADER sentiment scores.

| Detail | Features | Coeff. |
|---|---|---|
| Ridge regression with lags up to 3 | market_lag1 | 0.017038 |
| | market_lag2 | 0.008143 |
| | industry_lag2 | 0.006105 |
| | industry_lag1 | 0.005057 |
| | industry_lag3 | 0.004473 |
| Ridge regression with lags up to 5 | market_lag1 | 0.008773 |
| | market_lag3 | 0.001816 |
| | top_trending_products_lag1 | 0.001619 |
| | national_lag2 | 0.001597 |
| | magazines_lag4 | 0.001545 |

**Table 3**. Performance comparison of models with and without sentiment scores of news using DistilBERT sentiment scores.

| Model | RMSE (Without sentiment scores (lag 5)) | RMSE (With sentiment scores (lag 3)) | RMSE (With sentiment scores (lag 5)) |
|---|---|---|---|
| Linear Regression | 137.76 | **128.78** | 137.45 |
| Ridge Regression | 206.01 | **128.74** | 136.98 |
| Lasso Regression | 3372.02 | **223.45** | 224.52 |
| Elastic Net Regression | 175.88 | **133.79** | 135.77 |

From Table 1 and Table 3 it is quite evident that lagged values up to 3 lags are in a better position to predict the NIFTY 50 index movement. The models are capable of predicting the NIFTY 50 index value of only 1 day in future. The top 5 most positively contributing sentiment features are shown in Table 4 for the Ridge model since it was the best model found with DistilBERT based sentiment scores of news headlines.

**Table 4.** Top 5 positively contributing sentiment features in Ridge regression based on DistilBERT sentiment scores.

| Detail | Features | Coeff. |
|---|---|---|
| Ridge regression with lags up to 3 | market_lag3 | 0.026783 |
| | market_lag1 | 0.019582 |
| | market_lag2 | 0.018240 |
| | politics_and_nation_lag1 | 0.010176 |
| | international_lag3 | 0.009357 |
| Ridge regression with lags up to 5 | market_lag3 | 0.021886 |
| | politics_and_nation_lag5 | 0.017608 |
| | market_lag2 | 0.017419 |
| | market_lag1 | 0.017335 |
| | politics_and_nation_lag1 | 0.013276 |

Since the number of variables was quite high, feature importance plot having all features cannot be shown here. However, top 5 most positively contributing sentiment features are shown in Table 2 for Ridge Regression. The coefficients are quite low due to the fact that the lagged features of "Open", "High", "Low" and "Close" were all scaled down to the range of [0,1] for faster convergence. Hence, the target variable was also scaled down which led to smaller coefficient values. Ridge and Lasso Regression has hyperparameters and the hence, Grid Search algorithm was used to get the best set of hyperparameters. As per Table 4, two things have become quite prominent. Firstly, news related to market has greater amount of influence on the NIFTY 50 index and secondly, political and nation related news also have got good amount of influence on the movement of the said index.

V. CONCLUSION

The prime objective of the current study was to find out which news items have significant impact of the movement of NIFTY 50. In the present study sentiment scores were used to predict NIFTY 50 index of next day based on lagged values of the sentiment scores of news and news headlines respectively. The models used in this study were linear models namely Linear Regression, Ridge Regression, Lasso Regression and Elastic Net. The study showed that sentiment scores associated with different news articles have different level of impacts on the movement of index and it is not required to use more sophisticated ML based models to get a good prediction. The study focused on the impact of 22 different news topics on the movement of Nifty 50 and it was found out that news related to markets and politics have significant impact on the movement of this index. However, it is important to extract the sentiment scores as accurately as possible. The study showed that Ridge Regression performed in a much better way with sentiment scores extracted using DistilBERT. However, it would be wrong to generalize the findings for any other index or stock price. The current study focused only on the NIFTY 50 index. More comprehensive analysis can be done by considering multiple indexes and multiple stock prices also.

ACKNOWLEDGEMENTS

The researchers acknowledge the contributions of Debapratim Gupta, Ishika Sarkar, Aryan Dalal and Chandan Bhattacharya in this work by helping the researchers to collect the data based on the codes supplied by the researchers.